\documentclass[runningheads,a4paper]{llncs}

\usepackage{amssymb}
\usepackage{graphicx}
\usepackage{verbatim}

\usepackage{url}
\urldef{\mailsa}\path|{amit.gajbhiye, sardar.jaf, noura.al-moubayed, s.p.bradley}@durham.ac.uk|
\urldef{\mailsb}\path|{stephen.mcgough}@ncl.ac.uk|
\newcommand{\keywords}[1]{\par\addvspace\baselineskip
\noindent\keywordname\enspace\ignorespaces#1}

\begin{document}

\mainmatter  

\title{An Exploration of Dropout with RNNs for Natural Language Inference}

\titlerunning{An Exploration of Dropout with RNNs for Natural Language Inference}

%
%

\author{Amit Gajbhiye \inst{1}\and Sardar Jaf \inst{1}\and Noura Al Moubayed\inst{1}\and A. Stephen McGough\inst{2}\and Steven Bradley \inst{1} }
\authorrunning{An Exploration of Dropout with RNNs for Natural Language Inference}

\institute{Department of Computer Science, Durham University, Durham, UK\\
\mailsa\\
\and	
School of Computing, Newcastle University, Newcastle upon Tyne, UK
\mailsb\\
}

%
%

\toctitle{Lecture Notes in Computer Science}
\tocauthor{Authors' Instructions}

\maketitle

\begin{abstract}
Dropout is a crucial regularization technique for the Recurrent Neural Network (RNN) models of Natural Language Inference (NLI). However, dropout have not been evaluated for the effectiveness at different layers and dropout rates in NLI models. In this paper, we propose a RNN model for NLI and empirically evaluate the effect of applying dropout at different layers in the model. We also investigate the impact of varying dropout rates at these layers. Our empirical evaluation on a large (Stanford Natural Language Inference (SNLI)) and a small (SciTail) dataset suggest that dropout at each feed-forward connection severely affect the model accuracy at increasing dropout rates. We also show that regularizing the embedding layer is efficient for SNLI whereas regularizing the recurrent layer improves the accuracy for SciTail. Our model achieved an accuracy $86.14 \%$ on the SNLI dataset and $77.05 \%$ on SciTail.

\end{abstract}

\section{Introduction}

Natural Language Understanding (NLU) is the process to enable computers to understand the semantics of natural language text. The inherent complexities and ambiguities in natural language text make NLU challenging for computers. Natural Language Inference (NLI) is a fundamental step towards NLU \cite{maccartneyPhDThesis}. NLI involves logically inferring a hypothesis sentence from a given premise sentence.

The recent release of a large public dataset the Stanford Natural Language Inference (SNLI) \cite{snliData} has made it feasible to train complex neural network models for NLI. Recurrent Neural Networks (RNNs), particularly bidirectional LSTMs (BiLSTMs) 
have shown state-of-the-art results on the SNLI dataset \cite{drBiLSTMghaeini2018dr}. However, RNNs are susceptible to overfitting $-$ the case when a neural network learns the exact patterns present in the training data but fails to generalize to unseen data \cite{zaremba2014recurrent}. In NLI models, regularization techniques such as early stopping \cite{chen2017natural}, L2 regularization and dropout \cite{tayCompare} are used to prevent overfitting. 

For RNNs, dropout is an effective regularization technique \cite{zaremba2014recurrent}. The idea of dropout is to randomly omit computing units in a neural network during training but to keep all of them for testing. Dropout consists of element-wise multiplication of the neural network layer activations with a zero-one mask $(r_j)$ during training. Each element of the zero-one mask is drawn independently from $r_j  \sim \textrm{Bernoulli}(p)$, where $p$ is the probability with which the units are retained in the network. During testing, activations of the layer are multiplied by $p$ \cite{srivastava2014dropout}.

Dropout is a crucial regularization technique for NLI \cite{drBiLSTMghaeini2018dr}\cite{tayCompare}. However, the location of dropout varies considerably between NLI models and is based on trail-and-error experiments with different locations in the network. To the best of our knowledge no prior work has been performed to evaluate the effectiveness of dropout location and rates in the RNN NLI models.

In this paper, we study the effect of applying dropout  at different locations in an RNN model for NLI. We also investigate the effect of varying the dropout rate. Our results suggest that applying dropout for every feed forward connection, especially at higher dropout rates degrades the performance of RNN. Our best model achieves an accuracy of 86.14\% on the SNLI dataset and an accuracy of 77.05\% on SciTail dataset.

To the best of our knowledge this research is the first exploratory analysis of dropout for NLI. The main contributions of this paper are as follows:
(1) A RNN model based on BiLSTMs for NLI.
(2) A comparative analysis of different locations and dropout rates in the proposed RNN NLI model.
(3) Recommendations for the usage of dropout in the RNN models for NLI task.

The layout of the paper is as follows. In Section \ref{sec:relatedWork}, we describe the related work. In Section \ref{sec:rnnNLIModel}, we discuss the proposed RNN based NLI model. Experiments and the results are presented in Section \ref{sec:experiments}. Recommendations for the application of dropouts are presented in Section \ref{sec:recomendations}. We conclude in Section \ref{sec:conclusion}.

\section{Related Work} \label{sec:relatedWork}
The RNN NLI models follow a general architecture. It consists of : (1) an embedding layer that take as input the word embeddings of premise and hypothesis (2) a sentence encoding layer which is generally an RNN that generates representations of the input (3) an aggregation layer that combines the representations and; (4) a classifier layer that classifies the relationship (entailment, contradiction or neutral) between premise and hypothesis. 

Different NLI models apply dropout at different layers in general NLI architecture. NLI models proposed by Ghaeini et al. \cite{drBiLSTMghaeini2018dr} and Tay et al. \cite{tayCompare} apply dropout to each feed-forward layer in the network whereas others have applied dropout only to the final classifier layer \cite{innerAttentionLiu}. Bowman et al. \cite{snliData} apply dropout only to the input and output of sentence encoding layers.The models proposed by Bowman et al. \cite{bowman2016fast} and Choi et al. \cite{choi2017learning} applied dropout to the output of embedding layer and to the input and output of classifier layer. Chen et al. \cite{chen2017natural} and Cheng et al. \cite{cheng2016long} use dropout but they do not  elaborate on the location.

Dropout rates are also crucial for the NLI models \cite{munkhdalai2017neural}. Even the models which apply dropout at the same locations vary dropout rates.

Previous research on dropout for RNNs on the applications such as 
neural language models \cite{pachitariu2013regularization}, 
handwriting recognition \cite{pham2014dropout} and machine translation \cite{zaremba2014recurrent} have established that recurrent connection dropout should not be applied to RNNs as it affects the long term dependencies in sequential data.

Bluche et al. \cite{bluche2015apply} studied dropout at different places with respect to the LSTM units in the network proposed by \cite{pham2014dropout} for handwriting recognition. The results show that significant performance difference is observed when dropout is applied to distinct places. They concluded that applying dropout only after recurrent layers (as applied by Pham et al. \cite{pham2014dropout}) or between every feed-forward layer (as done by Zaremba et al. \cite{zaremba2014recurrent}) does not always yield good results. Cheng et al. \cite{cheng2017exploration}, investigated the effect of applying dropout in LSTMs. They randomly switch off the outputs of various gates of LSTM, achieving an optimal word error rate when dropout is applied to output, forget and input gates of the LSTM.

Evaluations in previous research were conducted on datasets with fewer samples. We evaluate the RNN model on a large, SNLI dataset (570,000 data sample) as well as on a smaller SciTail dataset (27,000 data samples).  Furthermore, previous studies concentrate only on the location of dropout in the network with fixed dropout rate.We further investigate the effect of varying dropout rates. We focus on the application of widely used conventional dropout \cite{srivastava2014dropout} to non-recurrent connection in RNNs.

\section{Recurrent Neural Network Model for NLI Task} \label{sec:rnnNLIModel}
\begin{figure}
	\centering
	\includegraphics[height=4cm]{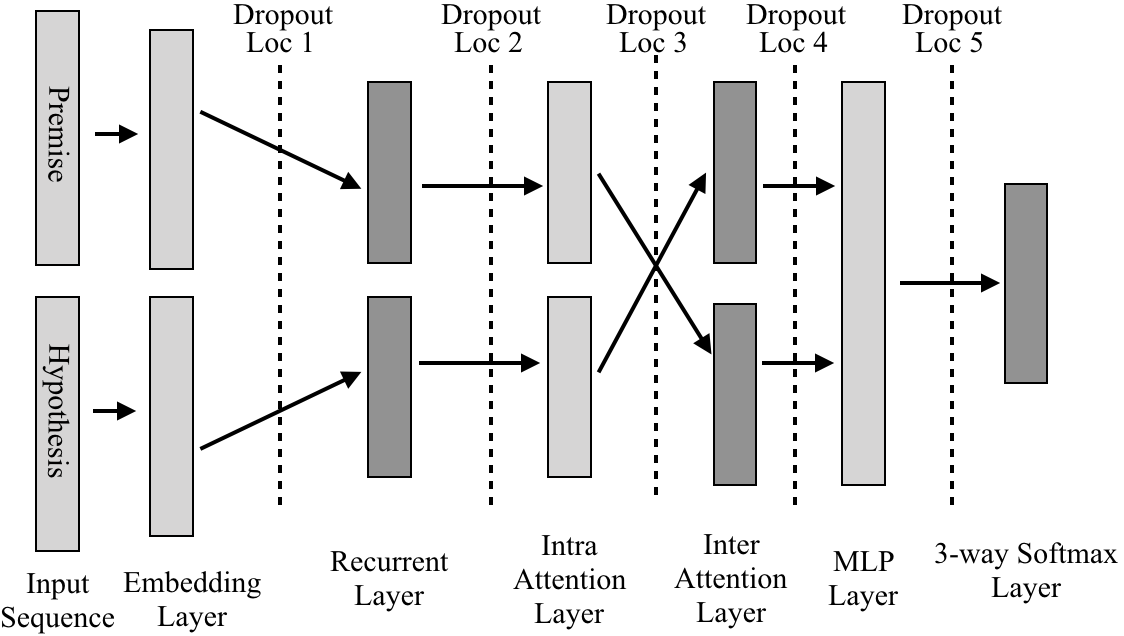}
	\caption{The Recurrent Neural Network Model with possible Dropout Locations }
	\label{fig: myNLIModel}
\end{figure}

The  proposed RNN NLI model follows the general architecture of NLI models and is depicted in Fig.\ref{fig: myNLIModel}. The model combines the intra-attention model \cite{innerAttentionLiu} with soft-attention mechanism \cite{kim2017neural}. 
The embedding layer takes as input word embeddings in the sentence of length $L$. The recurrent layer with BiLSTM units encodes the sentence. Next, the intra-attention layer generates the attention weighted sentence representation following the Equations $(\ref{eq:M})-(\ref{eq:R})$
\begin{equation} \label{eq:M}
M = \tanh\left(W^yY+W^hR_{avg}\otimes{e}_L\right)
\end{equation}
\begin{equation}\label{eq:alpha}
\alpha = softmax\left(w^TM\right)
\end{equation}
\begin{equation}\label{eq:R}
R = Y\alpha^T
\end{equation}
where, $W^y$, $W^h$ are trained projection matrices, $w^T$ is the transpose of trained parameter vector $w$, $Y$ is the matrix of hidden output vectors of the BiLSTM layer, $R_{avg}$ is obtained  from the average pooling of $Y$, $e_L \in {\mathbb{R} }^L$ is a vector of 1s, $\alpha$ is a vector of attention weights and $R$ is the attention weighted sequence representation. The attention weighted sequence representation is generated for premise and hypothesis and is denoted as $R_p$ and $R_h$. The attention weighted representation gives more importance to the words which are important to the semantics of the sequence and also captures its global context.

The interaction between $R_p$ and $R_h$ is performed by inter-attention layer, following the Equations $(\ref{eq:Iv}) - (\ref{eq:RhTilde})$.
\begin{equation}\label{eq:Iv}
I_v=R_p^TR_h
\end{equation}
\begin{equation}\label{eq:RpTilde}
\tilde{R_p}=softmax(I_v)R_h
\end{equation}
\begin{equation}\label{eq:RhTilde}
\tilde{R_h}=softmax(I_v)R_p
\end{equation}
where, $I_v$ is the interaction vector. $\tilde{R_p}$ contains the words which are relevant based on the content of sequence $R_h$. Similarly, $\tilde{R_h}$ contains words which are important with respect to the content of sequence $R_p$.
The final sequence encoding is obtained from the element-wise multiplication of intra-attention weighted representation and inter-attention weighted representation as follows:
\begin{equation}\label{eq:Fp}
F_p=\tilde{R_p}\odot{R_p}
\end{equation}
\begin{equation}\label{eq:Fh}
F_h=\tilde{R_h}\odot{R_h}
\end{equation}
To classify the relationship between premise and hypothesis a relation vector is formed from the encoding of premise and hypothesis generated in Equation (\ref{eq:Fp}) and (\ref{eq:Fh}), as follows:
\begin{equation}\label{eq:vpavg}
\textrm{v}_{p, avg}= \textrm{averagepooling} (F_p), \textrm{v}_{p, max}= \textrm{maxpooling}(F_p)
\end{equation}
\begin{equation}\label{vhavg}
\textrm{v}_{h, avg}= \textrm{averagepooling}(F_h), \textrm{v}_{h, max}= \textrm{maxpooling}(F_h)
\end{equation}
\begin{equation}\label{eq:Frelation}
F_{relation} = \left[\textrm{v}_{p, avg};\textrm{v}_{p, max};\textrm{v}_{h, avg};\textrm{v}_{h, max}\right] 
\end{equation}
where v is a vector of length $L$. The relation vector $\left(F_{relation}\right)$ is fed to the MLP layer. The three-way softmax layer outputs the probability for each class of NLI.

\section{Experiments and Results} \label{sec:experiments}
\subsection{Experimental Setup}
The standard train, validation and test splits of SNLI\cite {snliData} and SciTail \cite{scitail}  are used in empirical evaluations. The validation set is used for hyper-parameter tuning. The non-regularized model is our baseline model. The parameters for the baseline model are selected separately for SNLI and SciTail dataset by a grid search from the combination of L2 regularization $[1e-4, 1e-5, 1e-6]$, batch size $[32, 64, 256, 512]$ and learning rate $[0.001,0.0003, 0.0004]$. The Adam \cite{kingma2014adam} optimizer with first momentum is set to $0.9$ and the second to $0.999$ is used. The word embeddings are initialized with pre-trained 300-\textit{D} Glove 840\textit{B} vectors \cite{pennington2014glove}. Extensive experiments with dropout locations and hidden units were conducted however we show only the best results for brevity and space limits.

\subsection{Dropout at Different Layers for NLI Model}
Table \ref{T:diffModels} presents the models with different combinations of layers to the output of which dropout are applied in our model depicted in Fig. \ref{fig: myNLIModel}.
\begin{table}[!b]
	\centering
	\caption {Models with corresponding layers to the outputs of which dropout is applied.}
	\addtolength{\tabcolsep}{7 pt} 
	\renewcommand{\arraystretch}{1}
	\begin{tabular}{l l}
		\hline
		\textbf{Model} & \textbf{Layer}\\ \hline\hline
		Model 1 & No Dropout (Baseline)  \\ \hline
		Model 2 & Embedding  \\ \hline
		Model 3 & Recurrent  \\ \hline
		Model 4 & Embedding and Recurrent  \\ \hline
		Model 5 & Recurrent and Intra-Attention \\ \hline
		Model 6 & Inter-Attention and MLP  \\ \hline
		Model 7 & Recurrent, Inter-Attention and MLP  \\ \hline
		Model 8 & Embedding, Inter-Attention and MLP  \\ \hline
		Model 9 & Embedding, Recurrent, Inter-Attention and MLP \\ \hline
		Model 10 & Recurrent, Intra-Attention, Inter-Attention and MLP  \\ \hline
		Model 11 & Embedding, Intra-Attention, Inter-Attention and MLP  \\ \hline
		Model 12 & Embedding, Recurrent, Intra-Attention, Inter-Attention and MLP  \\ \hline
		Model 13 & Embedding, Recurrent, Inter-Attention and MLP  \\ \hline
	\end{tabular}
	\addtolength{\tabcolsep}{7 pt}
	\label{T:diffModels} 
\end{table}
Table \ref{T:modelAccuracy300U}. shows the results for the models in Table \ref{T:diffModels}. Each model is evaluated with dropout rates ranging from 0.1 to 0.5 with a granularity of 0.1.

\begin{table}[!t]
      \addtolength{\tabcolsep}{7 pt}
      \renewcommand{\arraystretch}{1.1} 
       \centering
		\caption{Model accuracy with varying dropout rates for SNLI and SciTail datasets. Bold numbers shows the highest accuracy for the model within the dropout range.}
		\begin{tabular}{ccccccc}
			\hline
			\textbf{Models} & \textbf{Dataset} & \multicolumn{5}{c}{\textbf{Dropout Rate (DR)}}
			\\ \hline \hline
			&  & 0.1 & 0.2 & 0.3 & 0.4 & 0.5 \\ \cline{3-7} 
			Model 1 & SNLI  & \multicolumn{5}{c}{84.45} \\ \cline{2-2}  \cline{3-7} 
			& SciTail & \multicolumn{5}{c}{74.18}  \\ \hline 
			
			Model 2 & SNLI  & 84.56 & 84.59 & 84.42 & \underline{\textbf{86.14}} & 84.85 \\ \cline{2-2}  \cline{3-7} 
			& SciTail & \textbf{75.45} & 75.12 & 74.22 & 73.10 & 74.08 \\ \hline
			
			Model 3 & SNLI  & 84.12 & \textbf{84.21} & 83.76&  81.04 & 79.63 \\ \cline{2-2}  \cline{3-7} 
			& SciTail & \underline{\textbf{76.15}} & 75.78 & 73.50 & 73.19 & 75.26 \\ \hline
			
			Model 4 & SNLI  & 83.83 & \textbf{85.22} & 84.34 &  80.82 & 79.92 \\ \cline{2-2}  \cline{3-7} 
			& SciTail & 74.65 & \textbf{76.08} & 74.22 & 74.46 & 73.19 \\ \hline
			
			Model 5 & SNLI  & \textbf{84.72} & 83.43 & 72.89 &  70.49 & 62.13 \\ \cline{2-2}  \cline{3-7} 
			& SciTail & \textbf{75.87} & 75.13 & 75.26 & 73.71 & 72.25 \\ \hline
			
			Model 6 & SNLI  & 84.17 & \textbf{84.32} & 83.71 &  82.79 & 81.68 \\ \cline{2-2}  \cline{3-7} 
			& SciTail & 73.85 & \textbf{75.68} & 75.26 & 73.95 & 73.28 \\ \hline
			
			Model 7 & SNLI  & \textbf{84.33} & 82.97 & 82.00 &  81.15 & 79.25 \\ \cline{2-2}  \cline{3-7} 
			& SciTail & 73.75 & \textbf{75.02} & 74.37 & 73.37 & 73.42 \\ \hline
			
			Model 8 & SNLI  & 84.67 & \underline{\textbf{85.82}} & 84.60 &  84.14 & 83.94 \\ \cline{2-2}  \cline{3-7} 
			& SciTail & 73.80 & 73.52 & 69.29 & \textbf{75.82} & 73.89 \\ \hline
			
			Model 9 & SNLI  & \textbf{84.44} & 83.05 & 82.09 &  81.64 & 79.62 \\ \cline{2-2}  \cline{3-7} 
			& SciTail & 75.68 & \underline{\textbf{76.11}} & 75.96 & 70.84 & 74.55 \\ \hline
			
			Model 10 & SNLI  & \textbf{84.45} & 80.95 & 75.31 &  70.81 & 69.34 \\ \cline{2-2}  \cline{3-7} 
			& SciTail & 73.30 & \textbf{75.21} & 74.98 & 74.65 & 71.59 \\ \hline
			
			Model 11 & SNLI  & \textbf{84.31} & 82.43 & 78.94 &  74.93 & 70.54 \\ \cline{2-2}  \cline{3-7} 
			& SciTail & \textbf{75.63} & 73.47 & 74.93 & 74.93 & 70.32 \\ \hline
			
			Model 12 & SNLI  & \textbf{84.32} & 82.60 & 73.36 & 71.53 & 66.67 \\ \cline{2-2}  \cline{3-7} 
			& SciTail & 73.47 & \textbf{75.63} & 74.74 & 73.42 & 74.40 \\ \hline
		\end{tabular}
	\addtolength{\tabcolsep}{7 pt}
	\label{T:modelAccuracy300U}
\end{table}

\textbf{Dropout at Individual Layers} We first apply dropout at each layer including the embedding layer. Although the embedding layer is the largest layer it is often not regularized for many language applications \cite{gal2016theoretically}. However, we observe the benefit of regularizing it. For SNLI, the highest accuracy is achieved when the embedding layer is regularized (Model 2, DR 0.4).

For SciTail, the highest accuracy is obtained when the recurrent layer is regularized (Model 3, DR 0.1). The dropout injected noise at lower layers prevents higher fully connected layers from overfitting. We further experimented regularizing higher fully connected layers (Intra-Attention, Inter-Attention, MLP) individually, however no significant performance gains observed.

\textbf{Dropout at Multiple Layers} 
We next explore the effect of applying dropout at multiple layers.  For SNLI and SciTail, the models achieve higher performance when dropout is applied to embedding and recurrent layer (Model 4, DR 0.2). This supports the importance of regularizing embedding and recurrent layer as shown for individual layers. 

It is interesting to note that regularizing the recurrent layer helps SciTail (Model 7, DR 0.2) whereas regularizing the embedding layer helps SNLI (Model 8, DR 0.2). A possible explanation to this is that for the smaller SciTail dataset the model can not afford to lose information in the input, whereas for the larger SNLI dataset the model has a chance to learn even with the loss of information in input. Also, the results from models 7 and 8 suggests that applying dropout at a single lower layer (Embedding or Recurrent; depending on the amount of training data) and to the inputs and outputs of MLP layer improves performance. 

We can infer from models 9, 10, 11 and 12 that applying dropout to each feed forward connection helps preventing the model overfit for SciTail (DR 0.1 and 0.2). However, for both the datasets with different dropout locations the performance of the model decreases as the dropout rate increases (Section \ref{sec:drEffect}).

\subsection{The Effectiveness of Dropout for Overfitting}
\begin{figure}[t]
\centering
\includegraphics[height=6cm, width=9.0cm]{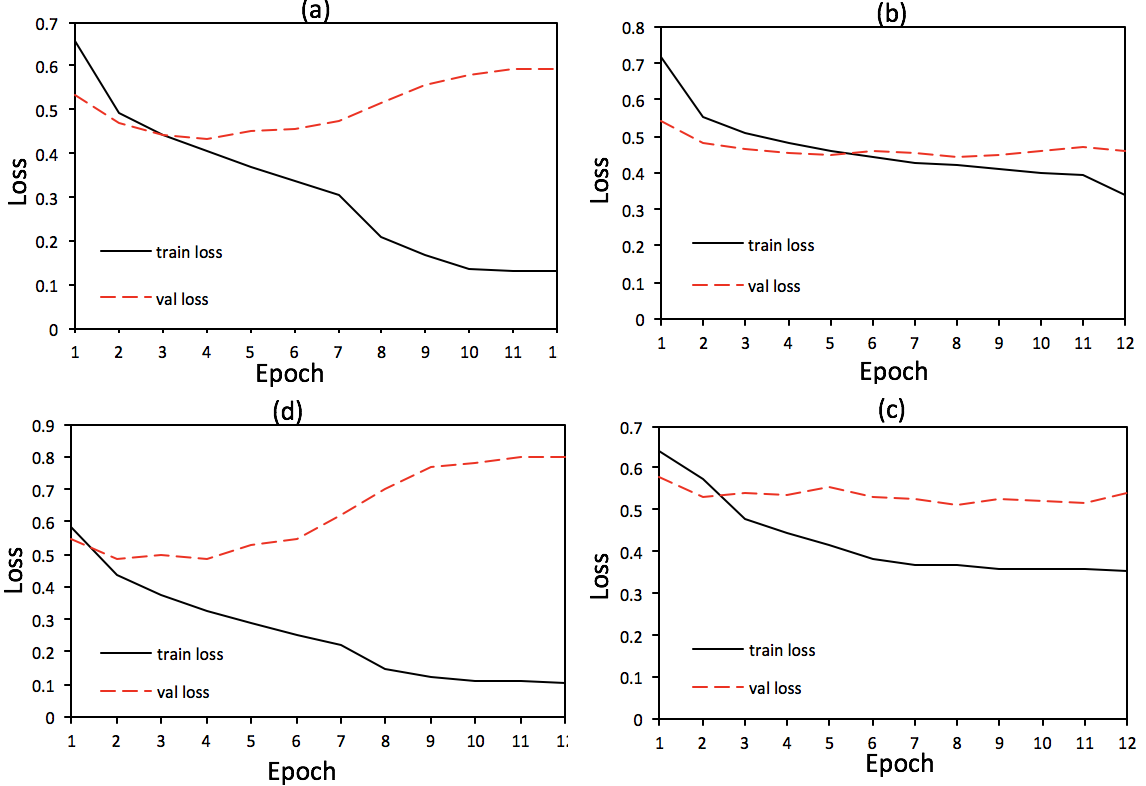}
\caption{Convergence Curves: (a) Baseline Model for SNLI, 
(b) Best Model for SNLI, 
(c) 100 Unit Model for SciTail, (d) 300 Unit Model for SciTail }
\label{fig:conCurv}
\end{figure}

We study the efficacy of dropout on overfitting. The main results are shown in Fig. \ref{fig:conCurv}. For SNLI, Fig. \ref{fig:conCurv} (a) - (b), shows the convergence curves for the baseline model and the model achieving the highest accuracy (Model 2, DR 0.4). The convergence curve show that dropout is very effective in preventing overfitting. However, for the smaller SciTail dataset when regularizing multiple layers we observe that the highest accuracy achieving model (Model 9, DP 0.2), overfits significantly (Fig. \ref{fig:conCurv}(d)). This overfitting is due to the large model size. With limited training data of SciTail, our model with higher number of hidden units learns the relationship between the premise and the hypothesis most accurately (Fig. \ref{fig:conCurv}(d)). However, these relationships are not representative of the validation set data and thus the model does not generalize well. When we reduced the model size (50, 100 and 200 hidden units) we achieved the best accuracy for SciTail at 100 hidden units (Table \ref{T:modelAcc100U}). The convergence curve (Fig. \ref{fig:conCurv}(c)) shows that dropout effectively prevents overfitting in the model with 100 hidden units in comparison to 300 units.
\begin{table}
\addtolength{\tabcolsep}{6 pt}
\renewcommand{\arraystretch}{1.1} 
\centering
\caption{Accuracy of 100 unit model for SciTail dataset}
	\begin{tabular}{ccccccc}
		\hline
		\textbf{Models} & \textbf{Dataset} & \multicolumn{5}{c}{\textbf{Dropout Rate (DR)}}
		\\ \hline \hline
		&  & 0.1 & 0.2 & 0.3 & 0.4 & 0.5 \\ \cline{3-7} 
		
		Model 13 &  SciTail & 76.72 & 76.25 & 72.58 & \underline{\textbf{77.05}} & 74.22 \\ \hline  
	\end{tabular}
\addtolength{\tabcolsep}{6 pt}
\label{T:modelAcc100U}
\end{table}
Furthermore, for SciTail dataset, the model with 100 units achieved higher accuracy for almost all the experiments when compared to models with 50, 200 and 300 hidden units.

The results of this experiment suggest that given the high learning capacity of RNNs an appropriate model size selection according to the amount of training data is essential. Dropout may independently be insufficient to prevent overfitting in such scenarios.

\subsection{Dropout Rate Effect on Accuracy and Dropout Location}
\label{sec:drEffect}
\begin{figure}
	\centering
	\includegraphics[height=5cm,width=8cm]{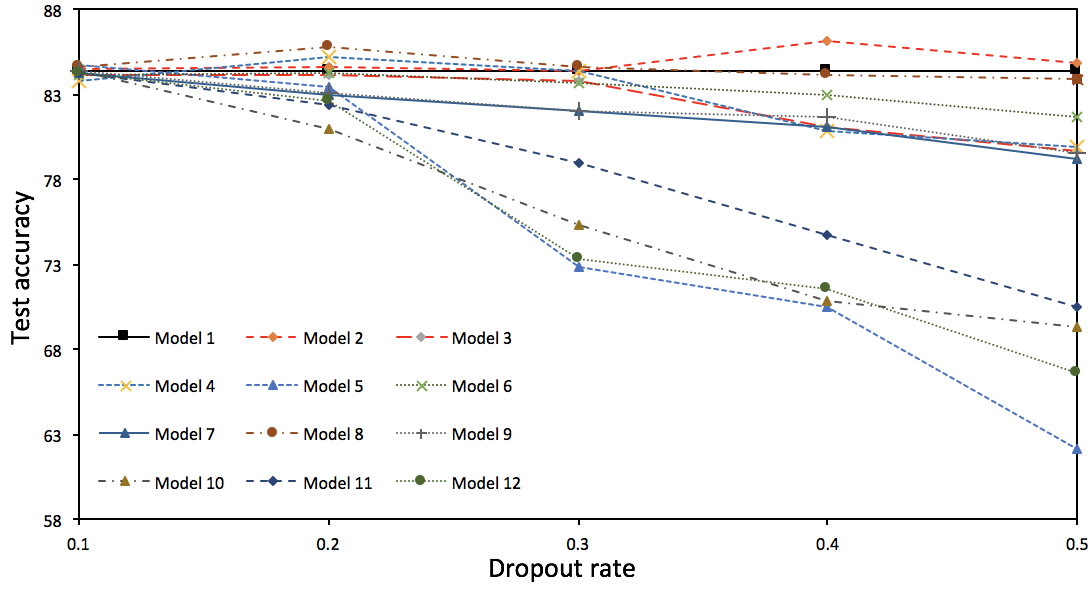}
	\caption{Plot showing the variation of test accuracy across the dropout range for SNLI.}
	\label{fig:accVsDR}
\end{figure}

We next investigate the effect of varying dropout rates on the accuracy of the models and on various dropout locations. Fig \ref{fig:accVsDR}. illustrates varying dropout rates and the corresponding test accuracy for SNLI. We observe some distinct  trends from the plot. First, the dropout rate and location does not affect the accuracy of the models 2 and 8 over the baseline. Second, in the dropout range [0.2 - 0.5], the dropout locations affect the accuracy of the models significantly. Increasing the dropout rate from 0.2 to 0.5 the accuracy of models 5 and 12 decreases significantly by 21.3\% and 15.9\% respectively.  For most of the models (3, 4, 6, 7, 9 and 10) the dropout rate of 0.5 decreases accuracy.

\begin{figure}
	\centering
	\includegraphics[height=5cm]{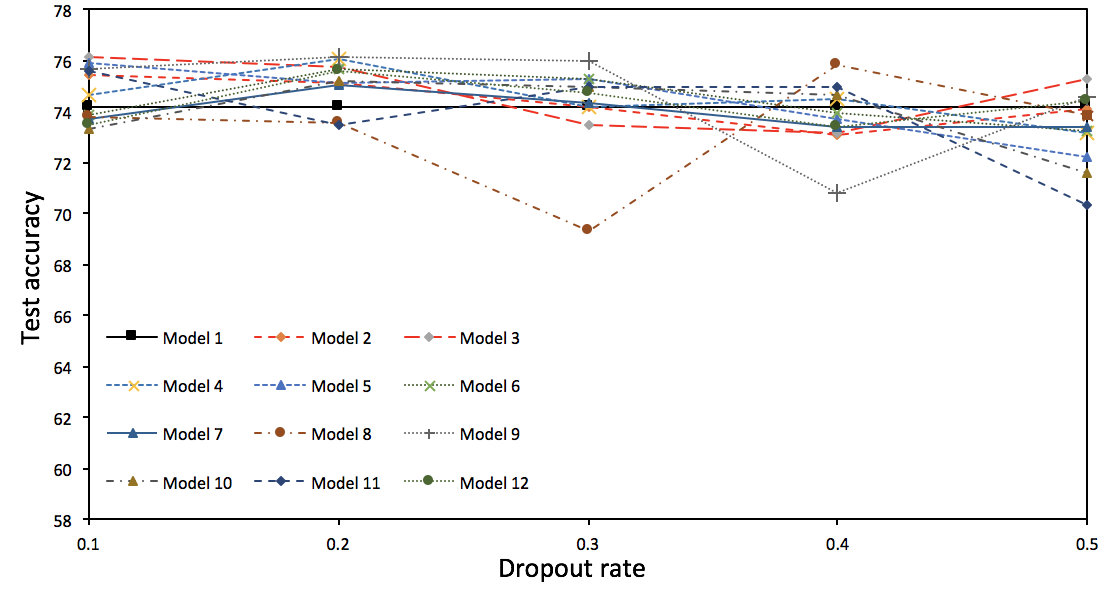}
	\caption{Plot showing the variation of test accuracy across the dropout range for SciTail.}
	\label{fig:SciaccVsDR}
\end{figure}

From the experiments on SciTail dataset (Fig. \ref{fig:SciaccVsDR}), we observed that the dropout rate and its location do not have significant effect on most of the models, with the exception of model 8 (which shows erratic performance).
Finally, for almost all the experiments a large dropout rate (0.5) decreases the accuracy of the models. The dropout rate of 0.5 works for a wide rang of neural networks and tasks \cite{srivastava2014dropout}. However, our results show that this is not desirable for RNN models of NLI. Based on our evaluations a dropout range of $[0.2 - 0.4]$ is advised.

\section{Recommendations for Dropout Application} \label{sec:recomendations}
Based on our empirical evaluations, the following is recommended for regularizing a RNN model for NLI task:
(1) Embedding layer should be regularized for large datasets like SNLI. For smaller datasets such as SciTail regularizing recurrent layer is an efficient option. The dropout injected noise at these layers prevent the higher fully connected layers from overfitting.
(2) When regularizing multiple layers, regularizing a lower layer (embedding or recurrent; depending on the amount of data) with the inputs and outputs of MLP layer should be considered. The performance of our model decreased when dropout is applied at each intermediate feed-forward connection.
(3) When dropout is applied at multiple feed forward connections, it is almost always better to apply it at lower rate  $-$ $[0.2 - 0.4]$.
(4) Given the high learning capacity of RNNs, an appropriate model size selection according to the amount of training data is essential. Dropout may independently be insufficient to prevent overfitting in the scenarios otherwise.

\section{Conclusions} \label{sec:conclusion}
In this paper, we reported the outcome of experiments conducted to investigate the effect of applying dropout at different layers in an RNN model for the NLI task. Based on our empirical evaluations we recommended the probable locations of dropouts to gain high performance on NLI task. 
Through extensive exploration, for the correct dropout location in our model, we achieved the accuracies of 86.14\% on SNLI and 77.05\% on SciTail datasets. In future research, we aim to investigate the effect of different dropout rates at distinct layers.

\medskip
\bibliographystyle{splncs03}
\bibliography{icann.bib}
\end{document}